# CLEAR: Cue Learning using Evolution for Accurate Recognition Applied to Sustainability Data Extraction

Peter J. Bentley
Department of Computer Science
UCL, Autodesk Research
London, United Kingdom
p.bentley@cs.ucl.ac.uk

Soo Ling Lim
Department of Computer Science
UCL
London, United Kingdom
s.lim@cs.ucl.ac.uk

Fuyuki Ishikawa
National Institute of Informatics
(NII)
Tokyo, Japan
f-ishikawa@nii.ac.jp

## ABSTRACT

Large Language Model (LLM) image recognition is a powerful tool for extracting data from images, but accuracy depends on providing sufficient cues in the prompt – requiring a domain expert for specialized tasks. We introduce Cue Learning using Evolution for Accurate Recognition (CLEAR), which uses a combination of LLMs and evolutionary computation to generate and optimize *cues* such that recognition of specialized features in images is improved. It achieves this by auto-generating a novel domain-specific representation and then using it to optimize suitable textual cues with a genetic algorithm. We apply CLEAR to the real-world task of identifying sustainability data from interior and exterior images of buildings. We investigate the effects of using a variable-length representation compared to fixed-length and show how LLM consistency can be improved by refactoring from categorical to real-valued estimates. We show that CLEAR enables higher accuracy compared to expert human recognition and human-authored prompts in every task with error rates improved by up to two orders of magnitude and an ablation study evincing solution concision.

## CCS CONCEPTS

• Computing methodologies ~ Machine learning ~ Machine learning approaches ~ Bio-inspired approaches ~ Genetic algorithms • Computing methodologies ~ Machine learning ~ Machine learning approaches ~ Neural networks

## 1 Introduction

Large Language Models (LLMs) are the leading approach today for image interpretation [1]. With appropriate prompts, LLMs trained on images and text can generate detailed textual descriptions of new images. This enables a new type of application: using LLMs to extract meaningful data from images. In e-commerce applications this might be to extract implicit attributes from a product image to improve search and user experience [2]. In medical applications this might be to describe the type of skin lesion an in attempt to identify melanoma early [3]. Here we focus on a new real-world application: the identification of sustainability data from images of buildings.

As the world becomes increasingly aware of the challenges presented by global warming, there is an urgent need to understand our existing buildings so that we can upgrade them to be more sustainable [4]. This is the challenge posed by our industrial partner, TheSqua.re Group – a global accommodation marketplace specializing in medium-term rentals all over the world. They have five data items of interest: *building age, lighting, heating, windows, energy*. Determining building age is often helpful to estimate likely construction methods and details such as insulation [5]. They must find structures that have inefficient lighting and heating so that they can be upgraded. They need to identify buildings with inefficient single glazed windows so that the latest thermally efficient double or triple glazing can be installed [6]. An overall estimation of energy kWh/m$^2$ enables the calculation of likely $CO_2$ emissions – often requested by corporate customers when booking accommodation.

In some European countries, modern buildings might have such information readily available for some, but not all properties [7]. For much of the world there is simply no available data at all about buildings [7]. However, while precise sustainability or energy related data may be sparse or non-existent, photographs and satellite imagery are plentiful. In the rental industry, it is the norm to take regular photographs not just to list the properties, but also to act as evidence against inventory loss or damages. Our industrial partner holds thousands of such images.

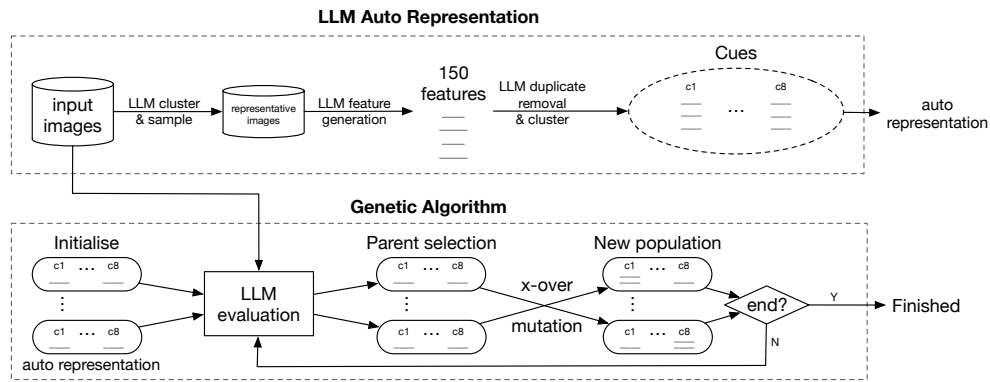

**Figure 1. Cue Learning using Evolution for Accurate Recognition (CLEAR) algorithm.**

Despite having image data available, the task of prompting an LLM to extract the required data from images is non-trivial [8]. One cannot simply prompt the LLM to list architectural features relating to energy usage – the answer is too generic and non-specific to be of value. Equally one cannot prompt the LLM to describe the type of heating or lighting from images – we receive non-specific and generic estimates as results. For this application we need the LLM to behave as a detective. It must collate several distinct features and use them to make a judgement. For example, an older building in Europe could be estimated if one considers cues such as ceiling height, coving, fireplaces, construction material visible as brickwork and size of windows. But these cues would be very different for images in other regions of the world that do not share such architectural traditions and practices.

So how best to determine the cues required in prompts? Our industrial partner needs an approach that does not depend on expert domain knowledge for each country, for this is not scalable and may not be available. All necessary information must be derived from the building images or general knowledge within an LLM. This paper proposes the solution: Cue Learning using Evolution for Accurate Recognition (CLEAR), see Figure 1. CLEAR uses a combination of LLMs and evolutionary computation to generate and optimize cues such that recognition of specialized features in images is improved.

This work makes the following contributions:
- Invention of the concept of *prompt cues* for LLMs.
- Introduction of CLEAR: a novel approach for improving the performance of LLMs by optimizing cues.
- Novel approach in EC that enables dynamic generation of representation for a genetic algorithm using LLMs.
- Application of CLEAR to a challenging real-world task.
- Demonstration of the effectiveness of CLEAR by comparison of our method with human-authored prompts and expert human evaluation.
- In-depth analysis of the algorithm, providing evidence that variable length operators and non-categorical encoding provide the best results.
- Demonstration that evolved solutions are coherent and concise through textual analysis and ablation studies.

The rest of the paper is organized as follows. We provide a background in the next section and give the method in Section 3. We describe empirical experiments in Section 4 and conclude in Section 5.

## 2 Background

The task of enhancing prompts is an area of extensive research, as generative systems are highly sensitive to exact wording, necessitating careful prompt engineering [9]. Techniques of improving prompts through text modifiers [10] or involving a human in the loop [11] are being developed.

Evolutionary approaches have been used with LLMs in various areas, such as game level generation [12] and neural architecture search [13]. Previous research has demonstrated how an LLM can optimize an objective function through prompt optimization [14], and how LLMs can be used for crossover operations [15]. Researchers have also demonstrated the use of LLMs as a means of initialization, crossover and mutation to evolve prompts that are then used to generate architectural images [16].

In the area of vision models, there is increasing use of vision models to understand and extract data, e.g., using LLMs for chart understanding by converting data-filled visuals into textual summaries [17] and previous work using LLMs to extract data from images of buildings [8], which demonstrated that the performance of LLMs in such scenarios are sensitive to the exact prompt used, hence justifying the approach in this paper.

The combination of EC with LLMs is becoming more common. Pluhacek et al. used LLMs to enhance the Self-Organizing Migrating Algorithm (SOMA), demonstrating that LLMs can be used to create distinctive and effective algorithmic strategies [18]. Morris et al. introduced the concept of "Guided Evolution" (GE), which is a novel framework that uses LLMs for a supervised evolutionary process, guiding mutations and crossovers [19], using the LLM to help maintain genetic diversity and augment decision-making in model evolution. Nasir et al.

proposed merging the code-generating abilities of LLMs with the diversity and robustness of Quality-Diversity (QD) solutions [20] to create diverse and high-performing networks. Liu et al. proposed an LLM-driven EA (LMEA), where each generation of the evolutionary search, LMEA instructs the LLM to select parent solutions from current population and perform crossover and mutation to generate offspring solutions [21]. EAs have also been used to explore the prompt space of LLMs, e.g., Saletta and Ferretti's grammar-based evolutionary approach [22] and Guo et al.'s EvoPrompt, which combines LLMs with EAs to automatically improve prompts [23].

## 3 Method

### 3.1 Overview

In this work we define a prompt **cue** to mean: *a signal or piece of information used to aid in the interpretation of an input by an LLM*. An example of a cue might be "window height", typically used within a prompt for the LLM in the form, "pay specific attention to window height."

CLEAR evolves cues in order to achieve accurate recognition of features by LLMs. It iteratively optimizes the behavior of an LLM by optimizing sets of textual cues. It makes use of a genetic algorithm with a dynamic LLM-generated representation, where each individual in the population is represented by a set $i$ of $n$ chromosomes, where $n$ is determined by an LLM:

$i = \{ ch_1, ch_2, ch_3, ..., ch_n \}$

and each chromosome $ch_x$ comprises a list of up to $m_x$ cues, where the allowable cues are also determined by an LLM:

$ch_x$ : [ $cue_1, cue_2, ..., cue_{mx}$ ]

Thus, each chromosome is a group of cues (grouped in terms of semantic or application-specific similarity), with its exact composition optimized by the GA. For example, the chromosome representing the set of "internal architectural features" might contain the cues "high ceilings" and "ceiling rose", with the prompt instructing the LLM to focus on the presence or absence of the cues high ceilings and ceiling rose.

We choose a genetic algorithm because LLM image interpretation is inconsistent with extremely noisy results – something GAs are particularly well-suited to handle.

While this work focusses on the real-world application of extracting sustainability data from images of buildings, CLEAR is a general technique that could be used to improve any prompt for LLM interpretation of an input. Figure 1 illustrates the technique.

### 3.3 Automated Representation

Before the GA can evolve cues, we must first generate the chromosomes and their associated allowable cues, i.e., we must create a representation for the GA to evolve. Traditionally this is crafted by hand. Here we introduce an automated approach using an LLM (gpt-4o). We do this to reduce bias caused by limited knowledge of a single expert and enable scalability.

**Table 1. Cue Generation LLM Prompts (*Windows*)**

| Purpose | LLM Prompt |
| --- | --- |
| Extract feature list | You are a surveyor. You are given a set of images that belong to the same building.<br>{ feature-extraction-prompt }<br>The building is located in UK. Return the features as a list. |
| { feature-extraction-prompt }: windows | Your task is to provide a detailed label of every architectural feature for the building that will help determine whether the glazing in the windows is single, double, or high efficiency. List 50 detailed visible features that are significant for window types. |
| Duplicate removal and clustering | I have a list of features: {raw_feature_list}. First, remove duplicated items, including features semantically similar. Then cluster these features based on the type of feature. Aim to produce 8 clusters. |
| Formatting | Given this list {categories}, first clean the list to contain text only, then produce a python array, each subarray for each category. |

We first cluster the image training set with respect to the data item of interest (using an LLM where necessary, e.g. for *Building Age* (see Supplementary materials for prompt details). We then select three representative buildings, randomly choosing one from each cluster. Next, the LLM is given each image separately and prompted to return a list of features, using a prompt specific to the data item. See Table 1 for an example prompt when the data item of interest is *windows* (Supplemental materials provide prompts for all data items). This results in 150 image features – our potential cues. We then ask the LLM to remove duplicates, cluster the features and format the output appropriately for our GA.

Our GA chromosomes are the categories created by the LLM during clustering. The possible values or genes within each chromosome are the cues within each cluster. For example, for windows we obtained $n=7$ chromosomes including $ch_0$ and $ch_2$:

*Material and Construction*, comprising $m_0=14$ cues including "Window Frame Material", "Window Thickness", "Frame Insulation", "Frame Color",…, "Depth of Window Frame"

*Functionality and Usability*, comprising $m_2=18$ cues including "Integrated Blinds/Shades", "Presence of Mullions or Muntins",…, "Glazing Pattern and Divisions" and so on.

This auto-representation using an LLM is able to produce many possibly useful cues for the optimizer to consider. Some are clearly derived from the images provided, some appear to be generated using its built-in general knowledge. However, the utility of them can be highly variable. For example, using the "Frame Color" cue from the chromosome *Material and Construction* is unlikely to help in the determination of energy efficiency. But using "Visual Consistency of the Glass Surface" instead might be useful – very old single glazed windows have distortions visible in their hand-made glass.

This illustrates the optimization problem for the genetic algorithm. With approximately eight chromosomes of up to 30 possible cues, which combination should be used in the prompt to maximize interpretation accuracy? It may be necessary to allow multiple cues per chromosome in case more than one in that category is important; equally we may need to allow zero cues in case that category is irrelevant.

## 3.4 Genetic Algorithms

We use two genetic algorithms: CLEAR$^x$ (fixed length), and CLEAR$^v$ (variable length), the latter having more complex crossover and mutation operators. This enables us to investigate whether CLEAR benefits from the ability to add more cues to the same category or even remove entire categories.

Our genetic algorithms manipulate cues (genes). Once the representation has been auto-generated, for every individual, for every chromosome, we initialize with a single random cue chosen from the corresponding LLM-generated cluster of possible cues.

To evaluate individual $i$, each $ch_x$ is extracted as a list of textual cues: *cue_list* and concatenated. These are then passed to the image interpretation LLM (gpt-4o) using the task-specific prompt (e.g., see Table 2). Chain-of-thought reasoning is encouraged by prompting the LLM to consider each cue in turn and give explanations before estimating. This is applied to each image. The interpreted result from the LLM is then cleaned (its textual output can be variable in formatting despite the prompt) and compared with the ground truth for each image to produce a fitness score (application-specific details in Section 4.2).

Our GAs permit elites. Should any individual in the population already have a fitness score, it is re-evaluated and the worst score is used for the individual. We also cache all past solutions and their fitnesses in a run; if a previously-seen combination of cues is recreated then it is also re-evaluated and the worst score amongst the cached copy and new score is used. This method, first used in evolutionary robotics [24], encourages solutions to be more consistent (robust against noise). Once all members of the population have been evaluated, we copy the best two into the next generation as elites. We then choose parents from the best $s$% of the population and generate the remainder of the next generation using crossover and mutation operators.

Crossover for the GA with fixed-length representation allocates a randomly chosen cue from either parent for each chromosome (uniform crossover). Mutation in a chromosome switches the cue with another randomly chosen from the same category. Thus, there is always the same number of chromosomes, each with $m_x$=1 cue.

Our crossover for the GA with variable-length representation operates as follows. Given parents *Parent1* and *Parent2*, we build each chromosome in turn for the child solution. Cues are iteratively randomly chosen from *Parent1*'s chromosome or *Parent2*'s chromosome until we reach the maximum number of cues of both parents. Should one parent have more cues in a chromosome, then the child has a probability of 0.5 of inheriting each additional cue in turn. Once complete, duplicate, or *overspecified* [25] cues are removed from the child. (If both parents share identical cues in a chromosome but in different orders then the child can be built with duplicates.) For example, if *Parent1* = { $ch_1$:[brick, concrete, wood] } and *Parent2* = { $ch_1$:[laminate, brick] } then one child could become *Child1* = { $ch_1$:[brick, brick] }, corrected to *Child1* = { $ch_1$:[brick] } while another could become *Child2* = { $ch_1$:[laminate, concrete, wood] }

Mutation for the variable-length representation can perform three operations, randomly chosen: swap, delete or add. First a random chromosome is chosen within the child solution. When swapping, mutation picks a random cue of that chromosome and swaps it with another allowable cue. If there is no cue in the chromosome, a new cue is added instead. When deleting, if a random cue can be chosen, it is deleted. When adding, another allowable cue is added to the chromosome. After mutation, duplicates are removed. For example, if both *Child1* and *Child2* were mutated they could become: *Child1* = { $ch_1$:[brick, steel] } and *Child2* = { $ch_1$:[laminate, concrete] }

Using these operators we permit the GA with variable-length representation to explore any possible combination of cues for each chromosome from zero to the full allowable set. Once the new population has been created, we re-evaluate and continue until the termination criteria is met (either a perfect fitness achieved, or the maximum number of generations has been reached), see Figure 1.

## 4 Experiments

Our experiments focus on the following four research questions:

R1. Validity of approach: Can cues be evolved from an LLM-generated representation that improve the accuracy of image interpretation?

R2. Assessment of algorithm: How does a GA with fixed-length representation compare with a GA using variable-length representation?

R3. Assessment of encoding: Can results be improved if categorical encoding is replaced with real encoding?

R4. Validity of solutions: Do the evolved sets of cues represent coherent, concise sets?

We first describe data preparation and application-specific details of fitness evaluation.

## 4.1 Data

This work is performed in collaboration with the global accommodation marketplace TheSqua.re. With their support we prepared a ground truth dataset of 47 apartments each with confirmed data (accurate sustainability data is rare). Our first step was the preparation of this dataset. For each of the 47 properties there were between 8 and 50 images provided. From these, four image subsets were chosen representing:

1. *Building images*: photographs of the exterior of the building if available, also wide-angle photographs showing as much of the interior as possible. (Used for *Building Age* and *Energy* data items)
2. *Heating images*: photographs that include objects that could be used for heating such as radiators, boilers or vents.
3. *Window images*: photographs that include windows, ideally showing features of frames, handles or other design characteristics.

4. *Lighting images*: photographs that include lights, including ceiling or wall mounted and free-standing, ideally showing details of bulbs.

While these were chosen manually in this work, the image selection stage could be automated by an LLM in future work. Our ground truth dataset also contains the true features contained by that apartment (laboriously collated from construction, owner, and surveyor documentation):

[D1] *Building age*, which may comprise: an exact year, e.g., 2014, a range, e.g., 2007-2011, or text depicting a time period, e.g., 19th century => cleaned to '1801-1900', before 1900 => cleaned to '1000-1899'.

[D2] *Lighting*, which comprises the percentage of lighting rated as low energy, e.g. 20% or 86%.

[D3] *Heating*, which comprises one of: underfloor heating, warm air, water radiators, electric panels or electric storage heaters.

[D4] *Windows*, comprising single glazed, double glazed, or high efficiency (triple glazed or high efficiency double glazing).

[D5] *Energy kWh/m²*: an integer typically between 35 and 450.

This dataset was split 60:40 for training and test respectively, ensuring that the distribution of data remained similar (in some cases there are very few samples, e.g. of very old buildings, so we ensure both training and test contain at least one).

## 4.2 Evaluation

For each of our five data items of interest: *Building age, Lighting, Heating, Windows* and *Energy kWh/m²* we have a corresponding evaluation function to compare the image interpretation LLM output with the ground truth. Supplemental material provides all prompts used; here we give an example of the prompts for Windows (Table 2). For *Building age,* the LLM returns a range. Eqn (1) is used to calculate error if there is an exact ground truth value for the age, while Eqn (2) is used if there is a range. For *Lighting*, the fitness score per building is simply the difference between estimated percentage of low energy and actual percentage Table 3 and Table 4 describe error calculations for *Heating* and *Lighting* respectively. For *Energy* the fitness is the absolute difference between estimated $kWh/m^2$ from the LLM and actual $kWh/m^2$ of the building. If the LLM estimates a range, Eqns (1) and (2) is used. Final fitness scores comprise the sum of errors per building in the dataset.

$$Error(startA, endA, pointB) = \begin{cases} 0, if\ (startA \leq pointB \leq endA) \\ \min\begin{pmatrix} |pointB - startA| \\ |pointB - endA| \end{pmatrix}, otherwise \end{cases} \quad (1)$$

$$Error(startA, endA, startB, endB) = \begin{cases} (startB - endA), if\ (endA < startB) \\ (startA - endB), if\ (endB < startA) \\ 0, otherwise \end{cases} \quad (2)$$

**Table 2. Cue Evaluation LLM Prompts for *Windows***

| Purpose | LLM Prompt |
|---|---|
| Evaluate cues | The images below belong to the same apartment. The building is located in UK.<br>{*prompt-question*}<br>Make your judgement focusing on the presence of the following features: {*cue_list*}<br>For each feature, say yes if it is visible, no if it is not visible or n/a if it is not applicable, then provide a short explanation.<br>{*instructions*}.<br>{*final-instructions*} |
| *prompt-question: windows* | "What type of windows does this apartment have?" |
| *instructions: windows* | "Finally, select one option: (1) single glazed, (2) double glazed, (3) high efficiency double or triple glazed" |
| *final-instructions* | "You can only use one of these, do not modify or invent your own options. Put the selected option in between ### and ###" |

**Table 3. *Heating* Error (LLM Estimate vs Ground Truth)**

|  | under floor | warm air | water rads | electric panel | electric storage |
|---|---|---|---|---|---|
| **underfloor** | 0 | 1 | 2 | 2 | 2 |
| **warm air** | 1 | 0 | 2 | 2 | 2 |
| **water rads** | 2 | 2 | 0 | 2 | 2 |
| **electric panel** | 2 | 2 | 2 | 0 | 1 |
| **electric storage** | 2 | 2 | 2 | 1 | 0 |

**Table 4. *Windows* Error (LLM Estimate vs Ground Truth)**

|  | Single glazed | Double glazed | High Efficiency |
|---|---|---|---|
| **Single glazed** | 0 | 1 | 2 |
| **Double glazed** | 1 | 0 | 1 |
| **High Efficiency** | 2 | 1 | 0 |

## 4.3 Experimental Setup

We perform four experiments to investigate our research questions:

- E1. Validity of approach. We test all five data items of interest: *Building age, Lighting, Heating, Windows* and *Energy kwh/m²* and compare the result from CLEAR with the result from using the same image recognition LLM using a prompt written by a domain expert, and with the result of using another human domain expert to perform the recognition task manually on the same images.
- E2. Assessment of operators. We test both versions of the GA: fixed length and variable length.
- E3. Assessment of encoding. We compare the performance of CLEAR using categorical vs real encodings.

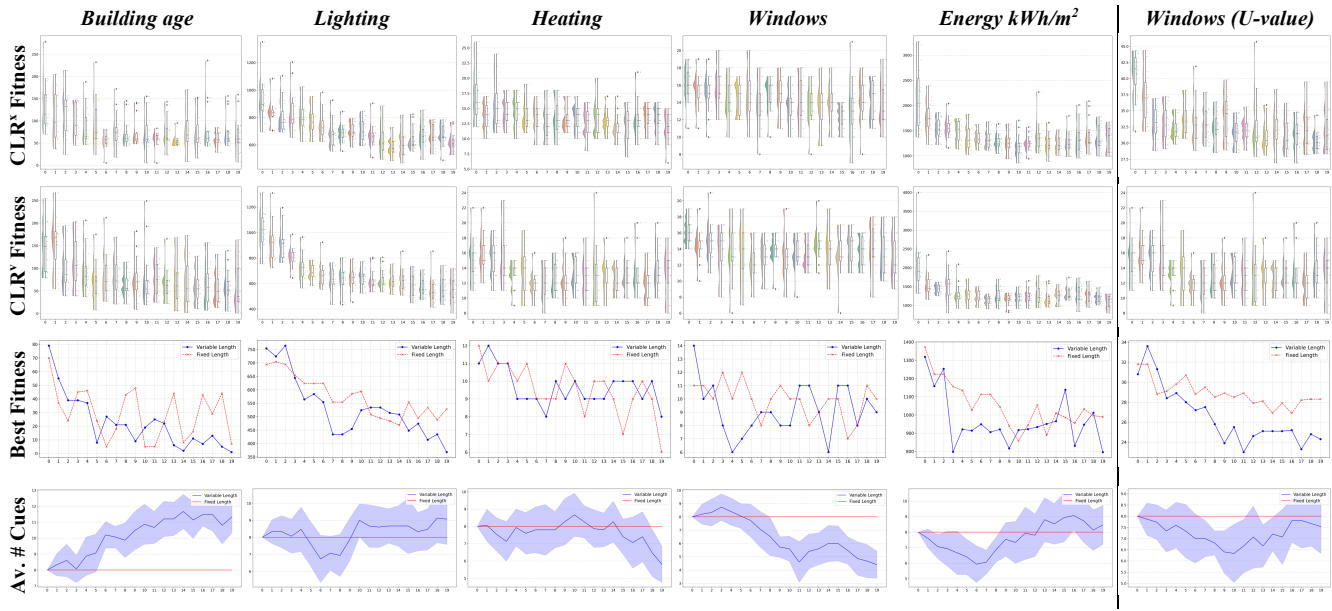

**Figure 2. Population fitness (raincloud plots), best fitness, and average number of cues (*y*-axis) over generations (*x*-axis), for each experiment. Final column shows result for experiment using U-Values for *Windows* in place of categorical values.**

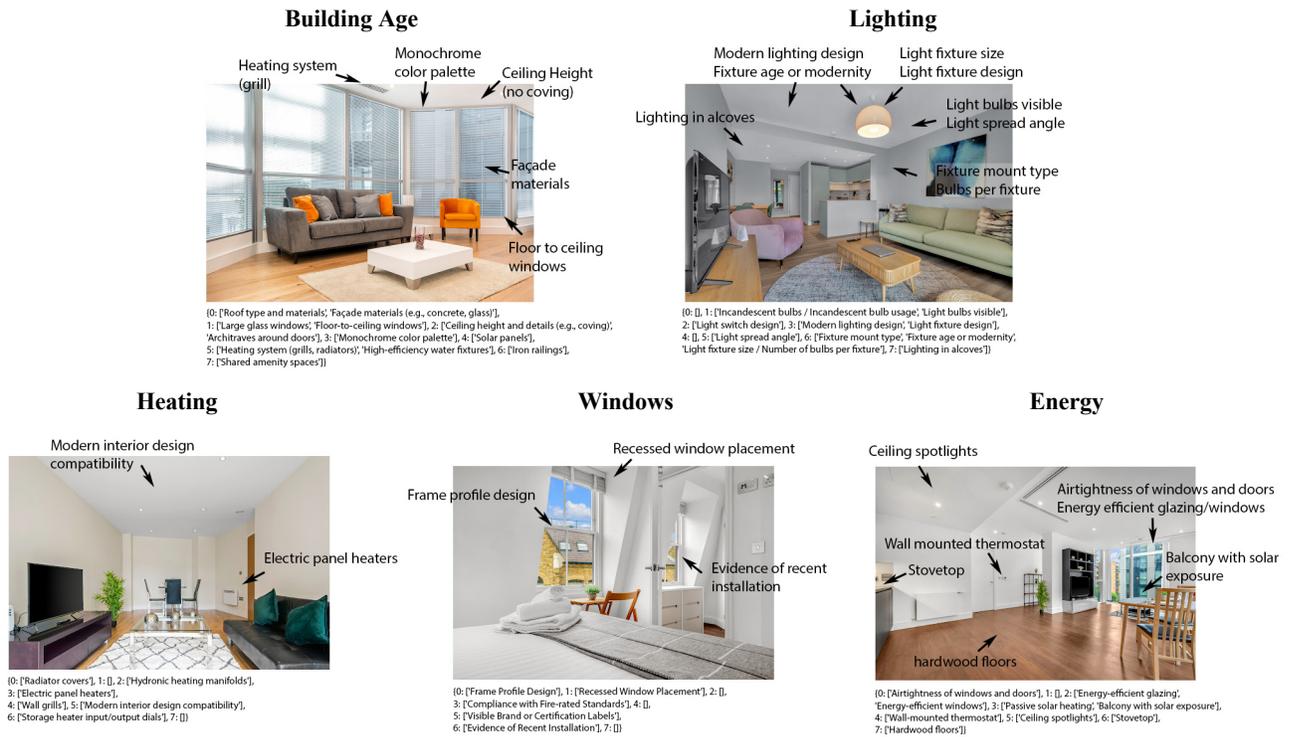

**Figure 3. Example images from the test dataset with cues of best evolved individuals by CLEAR$^v$ shown below. Relevant evolved cues highlighted for each data item of interest. (The LLM is presented with multiple images per dwelling to make its assessment.)**

**Table 5. Comparison of direct human image interpretation (Hu), expert-written prompt (Pr), and CLEAR using fixed representation (CLR$^x$) and variable-length representation (CLR$^v$). Lower values indicate better performance, best shown in bold, second best underlined. Ablation study shows new error for CLR$^v$ on test data after removal of each cue in turn, averaged.**

|  | Training error | | | | Testing error | | | | Ablation study | |
| --- | --- | --- | --- | --- | --- | --- | --- | --- | --- | --- |
|  | Hu | Pr | CLR$^x$ | CLR$^v$ | Hu | Pr | CLR$^x$ | CLR$^v$ | CLR$^v$ | σ |
| *Building age* | 147 | 86 | <u>7</u> | **1** | 129 | 49 | <u>20</u> | **9** | *18.9* | *19.99* |
| *Lighting* | 776 | <u>514</u> | 528 | **368** | 474 | <u>430</u> | 464 | **350** | *355* | *30.2* |
| *Heating* | 16 | 10 | **6** | <u>8</u> | 6 | 6 | 9 | 6 | *8* | *1.83* |
| *Windows* | 11 | 13 | <u>10</u> | **9** | 7 | 9 | 10 | **7** | *8.2* | *1.79* |
| *Energy kWh/m²* | 1311 | <u>959</u> | 990 | **796** | 984 | <u>916</u> | 1034 | **665** | *852* | *73.77* |

E4. Validity of solutions: We examine the cues in the final solutions to assess the quality of the cues with respect to the application. We also perform ablation studies, removing each cue in turn from the solution and observing if there is any change in error.

CLEAR$^x$ (fixed length), and CLEAR$^v$ (variable length) were both initialized with a population of 15 and run for 20 generations for each data item. *s*=33% of best parents were chosen as parents; the best two were duplicated into the new generation as elites. While larger population sizes, more generations and multiple runs are likely to produce superior results and enable statistical analysis, here we aim to prove the viability of the technique while using a modest time and compute budget (free LLM alternatives currently have insufficient quality output). As described in the previous section, LLM prompts are adjusted to construct relevant representations for each experiment.

## 4.4 Results

Table 5 and Figure 2 provide our results. Overall, for all experiments, CLEAR$^v$ outperforms or equals the performance of the other methods in both training and testing. CLEAR$^x$ can sometimes match the performance of direct human interpretation or expert-written prompts; in only one case it performs best (training dataset for *Heating*, but not for the test dataset). We revisit the research questions to understand the results in more detail.

*4.4.1 Validity of approach (R1).* CLEAR$^v$ (variable-length GA) successfully improves fitness scores over time. In several cases (*Age, Lighting, Energy*) its results are substantially better than direct human image interpretation and expert-written prompts (Table 5). It shows a clear pattern of changing the number of cues in chromosomes over time, occasionally removing all cues from some. Its approach varies according to the data item of interest. Runs where CLEAR$^v$ strongly outperforms other methods often show an evolution towards higher numbers of cues after an initial drop (Figure 2, bottom row). This may indicate initial pruning away of harmful or unneeded cues (and entire categories of cues) before adding new useful cues to the other chromosomes. This can be seen in the final solutions, which reveal that some chromosomes have zero cues while others have multiple (Figure 3). Results where CLEAR merely equals the performance of other methods show a reduction in cues, perhaps indictive of the noise and difficulty inherent in the problems: for some problems it may be easier to remove distractions that cause LLM variability than find useful new cues that improve accuracy.

*4.4.2. Assessment of operators (R2).* While CLEAR$^v$ successfully evolved high quality solutions, results for CLEAR$^x$ show far less coherence, see Table 5 and Figure 2. For most runs there is still evolutionary progression towards better accuracy, but (apart from *Heating*, training) final solutions are always worse compared to using a GA with variable-length representation. Final solutions still appear to contain cues that have limited utility. For this problem there appears to be a substantial advantage in having the ability to alter the number of cues.

*4.4.3 Assessment of encoding (R3).* CLEAR$^v$ works effectively for all data items of interest. However, it seems that it is more difficult to evolve cues for some data items compared to others. *Heating* and *Windows* show less improvement over time, with *Windows* showing a trend towards fewer cues over time, but some of the cues remaining still questionable, e.g. "Compliance with Fire-rated Standards" and "Visible Brand or Certification Labels" (none are visible). While CLEAR$^v$ equals or outperforms the other approaches, there remains a question: why is it easier to evolve cues for some data items compared to others?

Problem difficulty is likely an important factor, and only solvable with improved data and better LLMs. But another reason may be noise. The use of nondeterministic LLMs for image interpretation, combined with the difficult identification task, results in inconsistent results from the LLM for the same set of cues. The level of noise may also be correlated with the precise cues presented to the LLM. To understand this better we presented the LLM with the same single cue ten times for the same image. For cue "Evidence of Recent Installation" the mean difference in output for categories was 0.4 (40% were reported as a different type out of the three options) with coefficient of variation (CV) of 1.29. This is perhaps a relatively good cue, but inconsistently good – there is high variation. In contrast the same LLM with cue "Historical Building Integration" had difference of output 0.9, CV 0.35 – an example of a much worse cue, but less variation: it is more consistently worse.

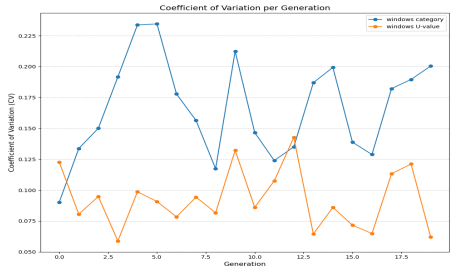

**Figure 4. Coefficient of Variation during evolution comparing categorical *Window* estimates and fitness (blue) with continuous U-value estimates and fitness (orange).**

It has been demonstrated in Bayesian Optimization of noisy problems that categorical and integer values (often via one-hot encoding or rounding) produce suboptimal results [26]. It is also evident that a stepped fitness function as we used for *Windows* (Table 4) will amplify variation – a minor error receives a large penalty – compared to a real-valued fitness function where a small error receives a correspondingly small penalty in fitness.

With this in mind, we modified our implementation for the Windows data item, asking it to estimate the real-valued U-value (a measure of insulator effectiveness) of the windows, low for efficient, high for inefficient. When we present the same single cues to the LLM interpreting the same image, now cue "Evidence of Recent Installation" produces a CV of 0.23 compared to cue "Historical Building Integration" which produces CV of 0.20. Now our unreliable cue has similar variation as the more reliable one. Pushing the change through to CLEAR and re-evolving cues for *Windows*, the new fitness function simply takes the difference in U-values between the LLM estimate and 0.5 (single glazed), 2.0 (double glazed) and 4.8 (efficient double or triple glazed). Evolution clearly shows a more stable optimization progression over time, with the number of cues no longer being minimized (Figure 2 rightmost column). Figure 4 shows the reduction in variation resulting from the change from categorical to real-valued estimates and shows a downwards trend suggesting that the GA can reduce this variation over time (caused by our approach of re-evaluating repeated individuals and always keeping the worst fitness scores, described earlier).

While final fitness scores cannot be directly compared because of the change in output and fitness function, this approach, which provides a much smoother gradient to follow, appears more promising for the optimizer. However, while it was possible to modify this data item, for *Heating* the different categories of heating cannot be so easily converted to a real-valued gradient.

4.4.4 *Validity of Evolved Solutions (R4)*. When we examine the evolved solutions from CLEAR$^v$, it is apparent that the evolved cues appear relevant and tailored to the data item in question. Taking *Energy kWh/m$^2$* as an example of an evolved solution: this data item is extremely difficult to estimate from photographs alone, even for domain experts. CLEAR performed better than the other approaches for both training and test datasets (Table 5) and showed a clear improvement over time (Figure 2) with the number of cues first falling as low as 5 or 6 and then increasing to around 10. Figure 3 (bottom right) shows the cues of the best individual from the final generation, which provides a useful set relating to energy efficiency, building design, energy-hungry appliances, modernity and quality of décor and presence of central heating. One category was discarded: "Heating and Ventilation" which contained seemingly useful cues such as "High-efficiency boiler", "Radiant floor heating", "Smart radiator valves", and "Underfloor heating controls". But such features are not visible in our dataset of images so they were unlikely to be useful for the LLM.

Relevance does not mean optimality, however. If CLEAR suffered from bloat then it is possible that our final solutions may contain many superfluous or unnecessary cues – or even cues that harm accuracy. To test for this, we performed ablation studies, removing each cue in turn from the five best solutions shown in Figure 3 and testing the new prompt with the image interpretation LLM to see if there is any change in error. Table 5 (far right) shows the results: in every case a removal of an evolved cue resulted in worse performance by the LLM when interpreting the images. It would appear that CLEAR evolves high quality and concise sets of cues.

## 5 Conclusions

The use of LLMs to derive useful data from photographs could be transformational as we strive to understand existing building stock to improve long-term sustainability. But an LLM must be given highly specific prompts in order to achieve accurate interpretation. With building styles and practices varied around the world, this makes the application of LLMs time-consuming and costly – local domain experts would need to engineer new prompts for every new region. Our task, set by the industrial partner, was to solve this accuracy and scalability problem.

We introduced an automated approach for specialized LLM image-to-data applications. Cue Learning using Evolution for Accurate Recognition (CLEAR) generates and optimizes textual *cues* such that the accuracy of deriving specific data (such as building age, heating type, energy kWh/m$^2$), for a given set of input data, is improved. LLMs optimized by CLEAR outperform human-authored prompts and expert human recognition by up to two orders of magnitude difference in error. Best results are achieved by using variable-length chromosomes to enable the number of cues to be changed over time. Real-valued estimates and corresponding fitness functions reduce noise and further improve accuracy compared to categorical estimates and stepped functions. An ablation study of final solutions provides evidence of concision, with the removal of any cue detrimental to accuracy.

This work has used image data of buildings in the UK. Future work will investigate the use of CLEAR for data from other regions. It is also possible that CLEAR may have utility for optimizing agentic LLM tasks and for other application domains.

## ACKNOWLEDGMENTS
We thank our industry partner TheSqua.re Group for providing their data and expertise for this research.

# CLEAR: Cue Learning using Evolution for Accurate Recognition Applied to Sustainability Data Extraction

Supplementary Materials

## 1 LLM Prompts to extract cues from images

> You are a surveyor. You are given a set of images that belong to the same building.
> {FEATURE_EXTRACTION_PROMPT}
> The building is located in UK. Return the features as a list.

| Data item | FEATURE_EXTRACTION_PROMPT |
|---|---|
| Building age | Your task is to provide a detailed label of every architectural feature for the building that will help determine the age of the building whether it is before 1900, 1900-1930, 1930-1950, 1950-1970, 1970-1990, 1990-2020, 2020-now. List 50 visible features that are significant for building age. |
| Lighting | Your task is to provide a detailed label of every visible feature in the images relating to artificial lights for the building that will help determine the type of lighting whether it is no low energy lighting, low energy in 20%, low energy in 40%, low energy in 60%, low energy in 80%, low energy in 100%. List 50 visible features that are significant for determining the type of bulbs used in the lights. Don't explain the label. |
| Heating | Your task is to provide a detailed label of every visible feature in the images relating to heating type that will help determine the type of heating used whether it is underfloor heating, water radiators, electric heaters, electric storage heaters or warm air from vents. List 50 visible features that are significant for determining the type of heating used in the apartment. Don't explain the label. |
| Windows | Your task is to provide a detailed label of every architectural feature for the building that will help determine whether the glazing in the windows is single, double, or high efficiency. List 50 detailed visible features that are significant for window types. |
| Energy kWh/m$^2$ | Your task is to provide a detailed label of every visible architectural feature, appliance and energy consuming device in the images that will help determine the energy consumption in kwh per metre squared. Do not list furnishings or belongings, focus on visible items relevant to energy consumption or saving. List 50, with no explanations. |

## 2 LLM Prompts to cluster training data

| Data item | Prompt |
|---|---|
| Building age | You are a surveyor. You are given this list of buildings, each row is a building with their id and the year they are built. First group the buildings by 3 eras to ensure good coverage representative of the architectural style and dataset, then return the ids of buildings per era in an array. |
| Lighting | N/A (no clustering required. Used 0%, 100%, anything in between) |
| Heating | N/A (no clustering required. Used water rads, electric panels, warm air) |
| Windows | N/A (no clustering required as there are only 3 values: single glazed, double glazed, and high efficiency) |
| Energy kWh/m$^2$ | N/A (no clustering required. Used <100, 100-200, >200) |

## 3 LLM Prompts to evaluate data item

> The images below belong to the same apartment. The building is located in UK.
> {PROMPT_QUESTION}
> Make your judgement focusing on the presence of the following features: {cue_list}
> For each feature, say yes if it is visible, no if it is not visible or n/a if it is not applicable, then provide a short explanation.
> {INSTRUCTIONS}.
> {FINAL_INSTRUCTIONS}

where {cue_list} is the list of cues generated by the Genetic Algorithm.

| Data Item | PROMPT_QUESTION | INSTRUCTIONS | FINAL_INSTRUCTIONS |
|---|---|---|---|
| Building age | What is the age of this apartment? | Finally, select one of these options: before 1900, 1900-1930, 1930-1950, 1950-1970, 1970-1990, 1990-2020, 2020-now | You can only use one of these, do not modify or invent your own options. Put the selected option in between ### and ### |
| Lighting | What type of lighting does this apartment have? | Finally, select one of these options: no low energy lighting, low energy in 20%, low energy in 40%, low energy in 60%, low energy in 80%, low energy in 100% | You can only use one of these, do not modify or invent your own options. Put the selected option in between ### and ### |
| Heating | What type of heating does this apartment have? | Finally, select one of these options: underfloor heating, water radiators, electric heaters, electric storage heaters, warm air from vents | You can only use one of these, do not modify or invent your own options. Put the selected option in between ### and ### |
| Windows | What type of windows does this apartment have? | Finally, select one of these options: (1) single glazed, (2) double glazed, (3) high efficiency double or triple glazed | You can only use one of these, do not modify or invent your own options. Put the selected option in between ### and ### |
| Energy kWh/m$^2$ | Estimate the energy consumption in kwh per metre squared for the following apartment. | Finally, give an estimate of the kwh. A highly efficient apartment might have a kwh/m2 value as low as 35 or better. An inefficient apartment might have a kwh/m2 value as high as 450 or worse | Put the estimated kwh in between ### and ###. Do not include any other text apart from the kwh values |

## 4 Algorithm Choices

It is the nature of real-world problems that external constraints affect choices, and this work is no exception. Through preliminary testing for this problem we found that only the biggest LLMs such as gpt-4o were able to parse our test images with any useful level of accuracy. Our collaborator also specifically requested the use of this model. With this constraint came time and budgetary constraints: experiments took many hours and cost several hundred dollars. (In rare cases, the LLM might produce unexpected outputs such that a run failed – we also implemented the ability to reload the population and continue the GA from that point, should this happen.) For these reasons it was necessary to use a simple GA with small population sizes of 15 with relatively few generations (20) as reported in the paper. For the same reasons it was not possible to perform each experiment multiple times in order to produce statistical analysis. Nevertheless, the reported findings were consistent throughout. With such a simple GA applied to a considerably noisy problem, hyperparameter values were straightforward: larger population sizes and longer runs produced better results, but our time and budget prohibited such options. Different mutation rates had little effect given the inherent noise produced by the LLM output. Thus, the paper studied those aspects which did have the largest effect, e.g., the encoding of the problem.

## 5 Source code

Source code is available here: https://github.com/soolinglim/clear/